\RequirePackage{fix-cm}
\documentclass[twocolumn]{svjour3}          
\smartqed  
\usepackage{graphicx}
\usepackage[ruled]{algorithm2e} 
\begin{document}

\title{Building Efficient CNN Architecture for Offline Handwritten Chinese Character Recognition}

\author{Zhiyuan Li         \and
        Nanjun Teng        \and
        Min Jin            \and
        Huaxiang Lu
}


\institute{F. Author \at
              first address \\
              Tel.: +123-45-678910\\
              Fax: +123-45-678910\\
              \email{fauthor@example.com}           
           \and
           S. Author \at
              second address
}

\institute{Z.Y. Li \at
              \email{lizhiyuan215@mails.ucas.ac.cn}           
             \and
             N.J. Teng \at
             \email{tengnanjun@semi.ac.cn}
	  \and
	  M. Jin \at
	 \email{jinmin08@semi.ac.cn}
	 \and
	 H.X. Lu \at
	  CAS Center for Excellence in Brain Science and Intelligence Technology \\
	  Beijing Key Laboratory of Semiconductor Neural Network Intelligent Sensing and Computing Technology \\
	\email{luhx@semi.ac.cn}
	\and
	Z.Y. Li, N.J. Teng, M. Jin, H.X. Lu \at
              Lab of Artificial Networks, Institute of Semiconductors, CAS \\
	   University of Chinese Academy of Sciences \\ }

\date{Received: date / Accepted: date}

\maketitle

\begin{abstract}
Deep convolutional neural networks based methods have brought great breakthrough in images classification, which provides an end-to-end solution for handwritten Chinese character recognition(HCCR) problem through learning discriminative features automatically. Nevertheless, state-of-the-art CNNs appear to incur huge computation cost, and require the storage of a large number of parameters especially in fully connected layers, which is difficult to deploy such networks into alternative hardware devices with limited computation amount. To solve the storage problem, we propose a novel technique called Global Weighted Arverage Pooling for reducing the parameters in fully connected layer without loss in accuracy. Besides, we implement a cascaded model in single CNN by adding mid output layer to complete recognition as early as possible, which reduces average inference time significantly. Experiments were performed on the ICDAR-2013 offline HCCR dataset, and it is found that our proposed approach only needs 6.9ms for classfying a chracter image on average, and achieves the state-of-the-art accuracy of 97.1\% while requires only 3.3MB for storage.
\keywords{Handwritten Chinese Character Recognition \and Convolutional Neural Networks \and Cascaded CNN}
\end{abstract}

\section{Introduction}
\label{intro}
Offline Handwritten Chinese Character Recognition has been a hot research filed over 40 years\cite{kimura1987modified} to deal with the challenges of large number of character classes, confusion between similar characters, and great diversity of handwritting style. In the last few years, a lot of traditional approaches have been proposed to improve recgnition accuracy but have yielded little progress. The study on better handcrafted features and strong classifiers have already reached their limit but the recognition acuracy is still far from huaman performance.

With the blooming growth of computational power, massive amounts of training data, and better nonlinear activation function, deep convolutional neural networks have achieved significant improvement in many computer vision tasks. After the application of CNNs in HCCR problem, the progress achieved in the past 5 years has greatly surpassed traditional approaches with large margin. However, deep convolutional neural networks are always invoved with billions of parameters and multiply-add accumulation. Such huge computation cost and storage requirements still prevents the use of CNN model in actual applications.

In this paper, we proposed an efficient CNN architecture for offline HCCR problem. Our network unifies the advantages of compact structure design, cascaded models and quantization. Firstly, we adopt the fire module proposed in Squeezenet\cite{iandola2016squeezenet} to design an extremely computation and storage efficient CNN architecture. Secondly, a new technical called Global Weighted Average Pooling was used to reduce the parameters of fully connected layers. Then we add extra mid output layers to classify most character images with about half computation cost of final output. At last, quantization was conducted at whole network for further compressing. We tested the proposed architecture on ICDAR-2013 offline competition database, and it is found 
that the it only takes 6.9ms for classfying a chracter image on average, while achieves a state-of-the-art accuracy of 97.1\% in the meanwhile.

The rest of this paper is organized as follows. Section \ref{Related Works} reviews the related works about Offline HCCR and CNN accelerating. Section \ref{Design of Compact Network} introduces the details about the proposed architecture, whereas Section \ref{Experiments} presents the experimental results, which include computation cost, run time and accuracy. The conclusions of this study and our future work are summarized in Section \ref{Conclusion}.

\section{Related Works}
\label{Related Works}

\subsection{Offline HCCR}

Handwritten Chinese Chracter Recognition(HCCR) has received intensive attention since early works in 1980s. The traditional HCCR approach involves three tasks: pre-processing, feature extraction and classification. Shape normalization is the most important part in pre-processing which can reduce the intra-class variations. Reseachers have proposed many useful shape normalization methods such as nonlinear normalization, bi-moment normalization, pseudo 2D normalization and line density projection interpolation\cite{liu2013online}. 
For HCCR, the structural features such as Gabor feature\cite{yong2002chinese} and gradient feature\cite{liu2007normalization-cooperated} will be more discriminable than structural features that mainly analyze the structure, strokes or parts of characters. The most commonly used models of classifiers include modiﬁed quadratic discriminant function(MQDF)\cite{kimura1987modified}, support vector machines(SVM)\cite{mangasarian2002data}, and discriminative learning quadratic discriminant function(DLQDF)\cite{liu2004discri}.

Due to the success of deep learning in image classification\cite{krizhevsky2012imagenet}\cite{simonyan2015very}\cite{szegedy2015going}\cite{he2016deep}, the research for offline HCCR has been changed from traditional methods to convolutional neural networks(CNN). Multi-column deep neural networks(MCDNN)\cite{ciregan2012multi-column}\cite{ciresan2015multi-column} was the first application of CNN for offline HCCR. MCDNN trained eight networks using different datasets. All networks have 4 convolutional layers and 2 fully connected layers. The best performance of single network achieves an accuracy of 94.47\%, which outperforms the best traditional method with significant gap. The accuracy is improved to 95.78\% with ensemble of 8 networks. In 2013 ICDAR offline competition\cite{yin2013icdar}, a research team from Fujistu developed a CNN-based method and took the winner place with an accuracy of 94.77\%. In 2014, they adopted a voting format of four alternately trained relaxation convolutional neural networks(ATR-CNN)\cite{wu2014handwritten}, which improved the accuracy to 96.06\%. Zhong et al.\cite{zhong2015high} combine the traditional feature extraction methods with character image as inputs, and adopted the inception architecture proposed in GoogLeNet. They reported an accuracy of 96.35\% , 96.64\%, 96.74\% with single, 4 and 10 ensemble models respectively, which became the first one beyond human performance. Zhang et al.\cite{zhang2017online} used the traditional gradient maps as network input, and obtained an accuracy of 96.95\%. With an additional adaptational layer, the recgnition accuracy can be improved to 97.37\%, which set new benchmarks for offline HCCR. Xiao et al.\cite{xiao2017building} adopted low-rank expansion and pruning technique to solve the problems of speed and storage capacity. The recognition of an character image took only 9.7ms on a CPU while required only 2.3MB for storage.
  
\subsection{Compressing and Accelerating}
In the early CNN structures, such as AlexNet\cite{krizhevsky2012imagenet} and VGG\cite{simonyan2015very}, the convolutional layers incur most of the computational cost and the fully connected layers contain the most network parameters.  In the last few years, there has been a significant amount of works on compresing the model size and accelerating inference speed. These works can be mainly divided into two categories: compressing pre-trained deep networks and designing compact layers.

\subsubsection{Compressing pre-trained deep network}
For fully connected layers, Chen et al.\cite{chen2015compressing} used the hash function to group weights into limited buckets, where connections in the same bucket share parameter value. Xue et al.\cite{xue2013restructuring} applied singular value decomposition(SVD) on the weight matrices in DNN, which reduce the parameters and accelerate the inference time on fully connected layers. Tensor factorization decompose weights into several pieces to compute convolutions with small kernels. Max et al.\cite{jaderberg2014speeding} constructed a low rank basis of filters that are rank-1 in the spatial domain. The 4D kernel matrix are decomposed into a combination of two 3D filters. Lebedev et al.\cite{lebedev2015speeding} used CP-decomposition to replace the original convolutional layer with a sequence of four convolutional layers with small kernels. Han et al.\cite{han2016deep} proposed deep compression which reduce the storage required by AlexNet from 240MB to 6.9MB by combining pruning, weight quantization, and Huffman coding. Some works\cite{li2016pruning}\cite{he2017channel} aimed to remove redundant channels on feature maps. Channels pruning does not result in sparse connectivity patterns and need special sparse convolution libraries like Han's method.

\subsubsection{Designing compact layers}
Lin et al.\cite{lin2014network} proposed a new strategy called global average pooling to replace fully connected layers with no parameters. Squeezenet\cite{iandola2016squeezenet} replaced $3\times3$ filters with $1\times1$ filters and decreased the number of input channels to $3\times3$ filters to design a very small network. The work in \cite{andrew2017mobilenets} introduced a depthwise separable convolution whcih only uses $8\sim9$ times less computation than standard convolutions. Pointwise group convolutions proposed in \cite{zhou2017shufflenet} reduces computation cost in dense $1\times1$ convolutions. BinaryNets\cite{courbariaux2016binarized}\cite{rastegari2016xnor:} constrained the weights and activations to $\pm1$ and replaced most floating-point multiplications by 1-bit exclusive-NOR operations. Although BinaryNets can provide the best compression rate and fastest speed-up, the accuracy drops a lot when dealing with large CNNs.

\section{Design of Efficient Network}
\label{Design of Compact Network}

\subsection{Fire Module}
A typical convolutional layer takes a $D\times D\times M$ feature map as input and produces a $D\times D \times N$ feature map where $D$ is the size of feature map, M is the number of input channels and N is the number of output channels. A typical convolutional layer has a trainable kernel $W$ of size $3\times 3 \times M \times N$ where $3$ is the most common spatial dimension of the kernel and $M$ is number of input channels and $N$ is the number of output channels. Standard convolutions have the computational cost of:
$9\cdot M\cdot N\cdot D\cdot D$. Forrest et al. \cite{iandola2016squeezenet} proposed a Fire module to reduce the parameters and computational cost in standard convolutional layer. A Fire module consists tow parts: squeeze layer which has only $1\times1$ filters, expand layer that has a mix of $1\times1$ and $3\times3$ convolutional filters. Squeeze layer produces a feature map with  $N/8$ channels, feeding into an expand layer which outputs two feature map with $N/2$ channels separately. A fire module has the computational cost of:

\begin{equation}
\frac{(M+5N)ND^2}{8}
\end{equation}

By replacing standard convolutional layer with Fire module, we get a reduction in computation( and storage) of:

\begin{equation}
\frac{1}{72}+\frac{5N}{72M}
\end{equation}

We adopt the fire module to design our baseline model, which uses $6\sim 9$ times less computation and storage than standard convolutional layer.

\begin{figure}[t]
\centering
\includegraphics[width=10pc]{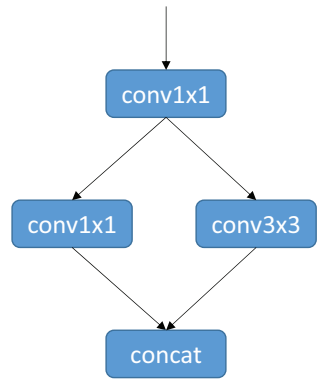}
\caption{Fire module: a squeeze convolution layer which has only $1\times1$ filters, feeding into an expand layer that has a mix of $1\times1$ and $3\times3$ filters.}
\label{fig_env1}
\end{figure}

\subsection{Global Weighted Average Pooling}
In early CNN architecture, convolution are performed in the lower layers of the network. For classification, the feature maps of the last convolutional layer are flattened and fed into fully connected layers. It treats the convolutional layers as feature extractors, and the extracted feature is classified with a traditional neural networks. In AlexNet and VGG-nets, extracted features are fed into three fully connected layers: the first two has 4096 neurons, the third performs 1000-way classification. This leads over 90\% parameters are storaged in fully connected layers. Instead of adding fully connected layers on the top of the feature maps, Lin et al. \cite{lin2014network} proposed Global Average Pooling to replace the traditional fully connected layers in CNN. They take the average of each feature map, and the result is fed directly into the softmax layer. Global average pooling just sums out the spatial information with equal weight, which leads worse result compared with additional fully connected layer in HCCR. To solve the accuracy drop, we add attention mechanism when performing global average pooling. An additional trainable kernel $W$ was added in this layer, and we call this strategy as weighted average pooling(WAP). Experiments showed that our method outperforms the global average pooling by a large margin with only a few parameters increased.
\begin{equation}
GWAP(x)=\sum_{jk}{w_{ijk}x_{ijk}}
\end{equation}

\begin{figure*}[!htb]
\centering
\includegraphics[width=40pc]{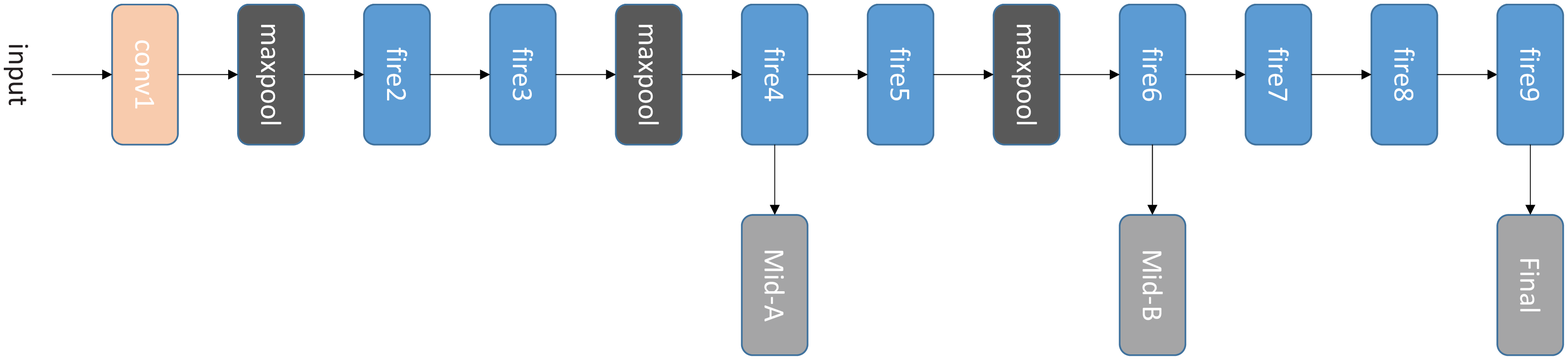}
\caption{Macroarchitectual view of our cascaded architecture.}
\label{fig_env1}
\end{figure*}

\begin{table*}[!htb]
\caption{Baseline model architectural dimensions}
\label{baselinemodelarchitecture}       
\centering
\begin{tabular}{ c  c  c  c  c  c  c  c  c }
\hline\noalign{\smallskip}
layer name & filter size/stride &output channels& depth & $s_{1\times1}$ & $e_{1\times1}$ & $e_{3\times3} $ & params & ops  \\
\noalign{\smallskip}\hline\noalign{\smallskip}
conv1 & $3\times3/1$ & 64 & 1 & & & & 832 & 2.4M \\
\noalign{\smallskip}\hline\noalign{\smallskip}
maxpool1 & $2\times2/2$ & 64 & 0 & & & &  &  \\
\noalign{\smallskip}\hline\noalign{\smallskip}
fire2 &  & 128 & 2 &16 &64 &64 &12K &11.5M \\
\noalign{\smallskip}\hline\noalign{\smallskip}
fire3 &  & 128 & 2 &16 &64 &64 &13K &12.6M \\
\noalign{\smallskip}\hline\noalign{\smallskip}
maxpool2 & $2\times2/2$ & 128 & 0 & & & &  &  \\
\noalign{\smallskip}\hline\noalign{\smallskip}
fire4 &  & 256 & 2 &32 &128 &128 &46K &11.5M \\
\noalign{\smallskip}\hline\noalign{\smallskip}
fire5 &  & 256 & 2 &32 &128 &128 &50K &12.6M \\
\noalign{\smallskip}\hline\noalign{\smallskip}
maxpool3 & $2\times2/2$ & 256 &  & & & &  &  \\
\noalign{\smallskip}\hline\noalign{\smallskip}
fire6 &  & 384 & 2 &48 &192 &192 &106K &6.7M \\
\noalign{\smallskip}\hline\noalign{\smallskip}
fire7 &  & 384 & 2 &48 &192 &192 &112K &7.0M \\
\noalign{\smallskip}\hline\noalign{\smallskip}
fire8 &  & 512 & 2 &64 &256 &256 &190K &12.1M \\
\noalign{\smallskip}\hline\noalign{\smallskip}
fire9 &  & 512 & 2 &64 &256 &256 &199K &12.6M \\
\noalign{\smallskip}\hline\noalign{\smallskip}
\noalign{\smallskip}
\end{tabular}
\end{table*}

\subsection{Cascaded Model in Single CNN}
Recalling the development of CNN structure used in HCCR, increasing the size of input character image and the layers of CNN model improved classification, but also need more computation.  The first successful CNN model applied in HCCR only has 4 convolutional layers and 2 fully connected layers, and achieved an accuracy of 94.47\% with 170 MFLOPs\cite{ciresan2015multi-column}. And the state-of-art CNN model has 12 layers, which achieve an accuracy of 97.59\% with 1.2GFLOPs\cite{xiao2017building}. It takes additional 6 times multiply-accumulation operations with 3.1\% accuracy increase. This inspired us to use shallow networks to recognition most easy character images with low computational cost, and the confusing character images are fed into a deeper network. The cascaded model complete most classification with shallow network, which is able to reduce the inference time on average. Further more, we implement the cascaded model by adding extra classifiers on the mid layers of a trained deep models. The mid output layers allows to complete classification with lower computational cost, and the final output ensures that the total model has a good performance. Besides, the average on different output results can increase the accuracy of final output. We add two mid  output layers: Mid-A and Mid-B. Mid-A is located after fire4, which consits a same fire module like fire5, a global weighted average pooling layer and a softmax layer for classification. Mid-B is located after fire6, follwed a same architecture like Mid-A, except the fire module has a same output like fire7.



\section{Experiments}
\label{Experiments}

\subsection{Experimental Data}
We used the offline CASIA-HWDB1.0(DB1.0) and CASIA-HWDB1.1(DB1.1) datasets for training our neural network, and evaluated our model on the ICDAR-2013 offline competition datasets. All datasets were collected by the Institute of Automation of the Chinese Academy of Sciences. The number of character classes is 3755(level-1 set of GB2312-80). Training dataset contains about 2.67 million samples contributed by 720 writers and competitionDB contains about 0.22 million samples contributed by another 60 writers\cite{liu2011casia}. 


\subsection{Experimental Settings}
Within a certain range, increasing the size of the input character image improved recognition accuracy, but leads higher computation cost. Considering the classification performance and the computation cost, we resized the input characters into $64\times64$. Since batch norm was added at all convolutional layers, we initilize the learning rate at 0.1, and then reduced it $\times0.1$ when the accuracy stop improving. We use stochastic gradient descent to train our model. The mini-batch is set to 256 with a momentum of 0.9.
To avoid overfitting, we add dropout layer before all softmax layer where the ratio was set to 0.5 for final output and 0.2 for mid output layers. Another regularization strategy we used is the weight decay with $L_2$ penalty. The multiplier for weight decay was set to $10^{-5}$ during the training process. We conducted the training process on Tensorflow\cite{abadi2016tensorflow:} using a GTX TITAN X GPU card.

\subsection{Comparision of HCCR-FC, HCCR-GAP and HCCR-WAP}
We designed 3 differnet architectures before performing classification, (1)We flattened the feature maps extracted by base network, and the resulting vector are fed into a hidden layer which has 1024 neurons(HCCR-FC); (2)We take the average of each feature map(HCCR-GAP); (3)We add  an additional trainable kernel $W$ to perform weighted average on each feature map(HCCR-WAP). We evaluated the performance comparison of different classification way on ICDAR-2013 competition dataset.  The recognition rate with HCCR-GAP we obtained was 96.47\%, whereas The recognition rate with HCCR-FC was 96.95\%. However, adding a fully connected layer before classification brings additional 33.6 millions parameters. The proposed weighted average pooling achieved a comparable accuracy of 96.91\% with only additional 0.03 millions parameters. It can be seen that replacing fully connected layers with global average pooling will cause significant decline on performance. And the proposed weighted average pooling can well balance the number of parameters and model accuracy.
Table \ref{Comparision_of_3_models} shows a comparison of these three models in classification accuracy and number of parameters.

\begin{table}[!htb]
\caption{Comparision of HCCR-FC, HCCR-GAP and HCCR-WAP}
\label{Comparision_of_3_models}       
\centering
\begin{tabular}{  c  c  c  }
\hline\noalign{\smallskip}
Model & Accuracy & Additional Params  \\
\noalign{\smallskip}\hline\noalign{\smallskip}
HCCR-GAP & 96.47\% & 0 \\
\noalign{\smallskip}\hline\noalign{\smallskip}
HCCR-FC & 96.95\% & 3.33M \\
\noalign{\smallskip}\hline\noalign{\smallskip}
HCCR-WAP & 96.91\% & 0.03M \\ \hline
\noalign{\smallskip}
\end{tabular}
\end{table}

\subsection{Training Strategy}
We trained the proposed architecture in two different ways:(1)Training all classifiers together; (2)Training the final classifiers firstly and fixed the parameters in base network, then training the extra mid output layers separatly. We found that multitask learning leads better performance at mid output layer but worse recognition rate at final classifiers. Compared with the state-of-the-art models, the performance of multi-task training drops a lot, thus we adopt the parameters trained by the second strategy to conduct the later experiments.
\begin{table}[!htb]
\caption{Comparision of Different Training Strategy}
\label{Comparision_of_training_strategy}       
\centering
\begin{tabular}{  c  c  c  c  }
\hline\noalign{\smallskip}
 & Mid-A & Mid-B & Final  \\
\noalign{\smallskip}\hline\noalign{\smallskip}
Multi-task Training & 95.13\% & 96.25\% & 96.31\% \\
\noalign{\smallskip}\hline\noalign{\smallskip}
Separatly Training & 94.36\% & 95.91\% & 96.91\% \\
\noalign{\smallskip}\hline\noalign{\smallskip}
\end{tabular}
\end{table}

\subsection{Results of Cascaded Model}
After training the model using DB1.0 and DB1.1, we measure the performance of middle output layers and the final output layer on each image of ICDAR-2013 offline competition datasets. We also measure the inference time on CPU by the middle output layers and the final output layer. The experiments are carried out on a PC equipped with 3.6GHz Intel Core i7-6700 and 8GB of memory. 

Table \ref{Comparision_of_mid_models} shows the experimental results including all output layers. Benefit from fire module, the final model achieves a state-of-the-art recognition rate with only 94MFLOPs. It is noteworthy that the inference speed for the final model is even faster than \cite{xiao2017building}. To furtherly save storage space for parameters, we use fixed point uniform quantification for the weights, and the impact of quantification can be negligible. For facilitate reading, we encode weights in 8bits for conv layer and 4 bits for fc layer.

\begin{table}[!htb]
\newcommand{\tabincell}[2]{\begin{tabular}{@{}#1@{}}#2\end{tabular}}
\caption{The running time for process a character image on CPU.}
\label{Comparision_of_mid_models}       
\centering
\begin{tabular}{  c  c  c  c  }

\hline\noalign{\smallskip}
Model & Accuracy &  \tabincell{c}{FLOPs\\($\times 10^8$)} & \tabincell{c}{ Inference\\ Time(ms) } \\
\noalign{\smallskip}\hline\noalign{\smallskip}
Mid-A & 94.36\% & 0.52 & 6.44\\
\noalign{\smallskip}\hline\noalign{\smallskip}
Mid-B & 95.91\% & 0.66 & 7.58\\
\noalign{\smallskip}\hline\noalign{\smallskip}
Final  & 96.91\% & 0.91 & 9.21\\
\noalign{\smallskip}\hline\noalign{\smallskip}
Cascaded  & 97.14\% & n/a & 6.93\\
\noalign{\smallskip}\hline\noalign{\smallskip}
CNN9Layer\cite{xiao2017building}  & 97.09\% & 1.52 & 9.77\\ \hline
\noalign{\smallskip}
\end{tabular}
\end{table}
 
 We use class probabilities as a good criteria for early inference. If the highest class probability exceeds a pre-defined threshold, the recognition is completed. It is no doubt that the bigger the threshold is, the samples will be classified with more computation, which means it will take more time to complete inference. Since the accuracy of Mid-B and final model is very close, we do not use Mid-B as middle output, but average the result of Mid-C and final model for better peroformance. We set the threshold as 0.98, and about 76\% of images can be classified at Mid-A which reduces the average inference time by 25\%. Besides, the use of Mid-B increases the accuracy by 0.2\%. 
\begin{table*}[!htb]
\centering
\caption{Different methods for ICDAR-2013 offline HCCR competition.}
\label{result comparison}       
\begin{tabular}{clclclclclc|}
\hline\noalign{\smallskip}
Method & Accuracy & Memory & Input & Representation Size & Ensemble \\
\noalign{\smallskip}\hline\noalign{\smallskip}
DFE + DLQDF\cite{liu2013online} & 92.72\% & 120.0MB & Gradient Feature & 512 & no \\
\noalign{\smallskip}\hline\noalign{\smallskip}
CNN-Fujitsu\cite{liu2013online} & 94.77\% & 2460.0MB & Gray Image & $1\times48\times48$ & yes(4) \\
\noalign{\smallskip}\hline\noalign{\smallskip}
MCDNN\cite{ciresan2015multi-column} & 95.79\% & 349.0MB & Gray Image & $1\times48\times48$ & no \\
\noalign{\smallskip}\hline\noalign{\smallskip}
ATR-CNN\cite{wu2014handwritten} & 96.06\% & 206.5MB & Binary Image & $1\times48\times48$ & yes(4) \\
\noalign{\smallskip}\hline\noalign{\smallskip}
Human Performance\cite{liu2013online} & 96.13\% & n/a & n/a & n/a & n/a \\
\noalign{\smallskip}\hline\noalign{\smallskip}
HCCR-Gabor-GoogLeNet\cite{zhong2015high} & 96.35\% & 27.77MB & Gabor Feature Maps + Gray Image & $9\times120\times120$ & no \\
\noalign{\smallskip}\hline\noalign{\smallskip}
DirectMap+ConvNet\cite{zhang2017online} & 96.95\% & 23.5MB & Direct Feature Maps & $8\times32\times32$ & no \\
\noalign{\smallskip}\hline\noalign{\smallskip}
HCCR-CNN9Layer\cite{xiao2017building} & 97.09\% & 2.3MB &  Gray Image & $1\times96\times96$ & no \\ 
\noalign{\smallskip}\hline\hline\noalign{\smallskip}
\bfseries Final output & \bfseries 96.91\% & \bfseries 10.3MB & \bfseries Gray Image & \bfseries $1\times64\times64$ & \bfseries no \\
\noalign{\smallskip}\hline\noalign{\smallskip}
\bfseries Cascaded Model& \bfseries 97.14\% & \bfseries 20.4MB & \bfseries Gray Image & \bfseries $1\times64\times64$ & \bfseries no \\
\noalign{\smallskip}\hline\noalign{\smallskip}
\bfseries Cascaded Model(Quantization) & \bfseries 97.11\% & \bfseries 3.3MB & \bfseries Gray Image & \bfseries $1\times64\times64$ & \bfseries no \\
\noalign{\smallskip}\hline
\end{tabular}
\end{table*}

\section{Conclusion and Future Work}
\label{Conclusion}
In this paper, we design an efficient CNN architecture for large scale HCCR involving 3755 classes. We proposed a weighted average pooling to balance the accuracy and the number of parameters. And we implemented a cascaded model in a sinle CNN by adding extra mid output layers, which reduce the average inference time significantly. These ideas can also be used in online HCCR. In future work, we plan to combine the proposed framework with other comprssing method to further reduce the model size.

\begin{acknowledgements}
This paper is supported by Beijing Technology Plan Project: Z171100002217094 and National Defense Science and Technology Project: 17-163-12-XJ-003-003-01.
\end{acknowledgements}



\end{document}